\documentclass[]{gmaart2022}

\renewcommand{\varProcName}{Proc. 33. Workshop Computational Intelligence}
\renewcommand{\varLocation}{Berlin}
\renewcommand{\varDate}{23.-24.11.2023}

\usepackage[utf8]{inputenc} 
\usepackage[T1]{fontenc} 
 
\usepackage{amsmath,amssymb,amsfonts,mathtools,amsthm} 

\usepackage[babel,autostyle]{csquotes}

\usepackage{tabularx}
\newcolumntype{L}[1]{>{\raggedright\arraybackslash}p{#1}} 
\newcolumntype{C}[1]{>{\centering\arraybackslash}p{#1}} 
\newcolumntype{R}[1]{>{\raggedleft\arraybackslash}p{#1}} 
\usepackage{booktabs}

\usepackage{paralist}   

\usepackage[noend]{algpseudocode}
\newcommand*\Let[2]{\State #1 $\gets$ #2}
\algrenewcommand\algorithmicrequire{\textbf{Voraussetzung:}}
\algrenewcommand\algorithmicensure{\textbf{Abschlussbedingung:}}

\usepackage{listings} 
\usepackage{subcaption}  

\usepackage{epstopdf}
\usepackage{xcolor}
\selectcolormodel{gray}  

\usepackage{scrhack}
\usepackage{varwidth}
\usepackage{rotating} 
\usepackage[defaultlines=2,all]{nowidow}  
\usepackage{import}
\usepackage{placeins}
\usepackage{makecell}


\usepackage{amsmath,amssymb,amsfonts,mathtools,amsthm} 

\usepackage{longtable} 
\usepackage{multirow}
\usepackage{tikz}
\usepackage{pgfplots}
\usepackage{smartdiagram}
\usetikzlibrary{patterns}
\usetikzlibrary{shapes} 
\usetikzlibrary{shapes.geometric, arrows,positioning}
\usepackage{smartdiagram}
\usesmartdiagramlibrary{additions} 
\AtBeginDocument{
}
\usepackage{xcolor}
\selectcolormodel{gray}

\usepackage[acronym]{glossaries} 
\usepackage{hyphenat}
\usepackage{etoolbox}
\usepackage{printlen}
\usepackage{graphicx}
\usetikzlibrary{calc, positioning, matrix}
\usepackage{standalone}

\newcommand{\newacr}[4][]{\newacronym[
	sort={\ifthenelse{\isempty{#1}}{#2}{#1}},
	]{#2}{#3}{#4}}
\glsdisablehyper

\newcommand{\etal}{et al. \@}

\robustify{\bfseries}

\usepackage{comment}
\usepackage{glossaries}

\newacronym{mlops}{MLOps}{Machine Learning Operations}
\newacronym{ai}{AI}{Artificial Intelligence}
\newacronym{devops}{DevOps}{Development and Operations}
\newacronym{ml}{ML}{Machine Learning}
\newacronym{dl}{DL}{Deep Learning}
\newacronym{ci}{CI}{Continuous Integration}
\newacronym{cd}{CD}{Continuous Deployment}
\newacronym{ssl}{SSL}{Self-Supervised Learning}
\newacronym{ct}{CT}{Continuous Training}
\newacronym{iac}{IaC}{Infrastructure as Code}
\newacronym{automl}{AutoML}{Automated Machine Learning}
\newacronym{dac}{DAC}{Dynamic Algorithm Configuration}
\newacronym{hpo}{HPO}{Hyperparameter Optimization}
\newacronym{nas}{NAS}{Neural Architecture Search}
\newacronym{ae}{AE}{autoencoder}
\newacronym{mae}{MAE}{masked autoencoder}
\newacronym{cash}{CASH}{Combined Algorithms Selection and Hyperparameter Optimization}
\newacronym{rest}{REST}{Representational State Transfer}
\newacronym{mse}{MSE}{Mean Squared Error}
\newacronym{nacip}{NACIP}{Natural, Artificial and Cognitive Information Processing}
\newacronym{haicore}{HAICORE}{Helmholtz AI COmpute REssources}
\newacronym{hai}{Helmholtz AI}{Helmholtz Artificial Intelligence Cooperation Unit}
\newacronym{gan}{GAN}{Generative Adversarial Network}
\newacronym{vae}{VAE}{Variational Autoencoder}
\newacronym{pomlia}{POMLIA}{Production-Oriented Machine Learning for Image Analysis}

\begin{document}


\hyphenpenalty=2000

\pagenumbering{roman}
\cleardoublepage
\setcounter{page}{1}
\pagestyle{scrheadings}
\pagenumbering{arabic}

\setnowidow[2]
\setnoclub[2]

\renewcommand{\Title}{MLOps for Scarce Image Data: A Use Case in Microscopic Image Analysis}

\renewcommand{\Authors}{
    Angelo Yamachui Sitcheu$^{1}$, Nils Friederich$^{1,2}$, Simon Baeuerle$^{1,3}$, Oliver Neumann$^{1}$, Markus Reischl$^{1}$, Ralf Mikut$^{1}$
}
\renewcommand{\Affiliations}{
    $^{1}$Institute for Automation and Applied Informatics, \\
    $^{2}$Institute of Biological and Chemical Systems, \\
    Karlsruhe Institute of Technology,
    Hermann-von-Helmholtz-Platz 1, \\
    76344 Eggenstein-Leopoldshafen \\
    E-Mail: angelo.sitcheu@kit.edu
    \\
    $^{3}$Robert Bosch GmbH,\\
    Markwiesenstraße 46, 72770 Reutlingen
}

\renewcommand{\AuthorsTOC}{A. Yamachui Sitcheu, N. Friederich, S. Baeuerle, O. Neumann, M. Reischl, R. Mikut} 
\renewcommand{\AffiliationsTOC}{Karlsruhe Institute of Technology, Robert Bosch GmbH} 

\setLanguageEnglish
							 
\setupPaper 


\section*{Abstract}
\label{sec:abstract}
Nowadays, \acrfull{ml} is experiencing tremendous popularity that has never been seen before. The operationalization of \acrshort{ml} models is governed by a set of concepts and methods referred to as \acrfull{mlops}. Nevertheless, researchers, as well as professionals, often focus more on the automation aspect and neglect the continuous deployment and monitoring aspects of \acrshort{mlops}. As a result, there is a lack of continuous learning through the flow of feedback from production to development, causing unexpected model deterioration over time due to concept drifts, particularly when dealing with scarce data. This work explores the complete application of \acrshort{mlops} in the context of scarce data analysis. The paper proposes a new holistic approach to enhance biomedical image analysis. Our method includes: a fingerprinting process that enables selecting the best models, datasets, and model development strategy relative to the image analysis task at hand; an automated model development stage; and a continuous deployment and monitoring process to ensure continuous learning. For preliminary results, we perform a proof of concept for fingerprinting in microscopic image datasets.

\section{Introduction}
\label{sec:introduction}
In the field of image analysis, \acrfull{ml}, particularly its subfield \acrfull{dl} is being explored to model complex problems such as image registration, classification, segmentation, object detection and tracking \cite{chan2020deep, DBLP:journals/ijmir/SuganyadeviS022}. In this context, the goal of \acrshort{ml} is to build a model that generalizes across different images for the same analysis task, for example, image segmentation. Therefore, the research community focuses more on developing new methods that achieve better performances and are computationally more efficient \cite{mavska2023cell}.
While developing the \acrshort{ml} model offline seems easy and cheap, operationalizing the model, which means, deploying the model and maintaining its performance over time, still faces numerous challenges \cite{Hechler2020}. Unlike standard non \acrshort{ml}-based software whose operations include building, testing, deployment and monitoring, \acrshort{ml} systems are more complex due to the two new components model and data, and their direct, and generally challenging relationship. These systems are often a source of high technical debt when not operationalized consequently \cite{DBLP:conf/nips/SculleyHGDPECYC15}. Similarly to standard non \acrshort{ml}-based software whose lifecycle follows the \acrfull{devops} scheme \cite{devopsgrundlage}, \acrshort{ml} systems have their development and operation paradigm known as \acrfull{mlops}(see Section \ref{subsec:mlops}).

Nevertheless, as shown by the multitude of approaches developed \cite{mavska2023cell}, it is very challenging to develop and operationalize one single model that generalizes across different images for the same task. Many experts develop new models for the same image analysis task when new input data is available. Due to the time and cost expensiveness of this process, they often focus more on developing the model and neglect the continuous monitoring and deployment aspect \cite{DBLP:journals/access/KreuzbergerKH23}, resulting in a lack of continuous learning through the flow of feedback from production to development, and therefore to unexpected model deterioration over time \cite{DBLP:conf/nips/SculleyHGDPECYC15}. Additionally, the initial training data in this field is often scarcity-prone in quantity and quality due to the tediousness of several tasks such as image acquisition and image annotation requiring skilled and expensive experts. This often leads to a dataset containing less relevant data from the experiments and more noise.

A potential solution is to create a production-oriented machine learning approach that harnesses the full range of available and well-established datasets and models to improve the efficiency of image analysis across multiple tasks.\\
This will be the main research goal of our work. In our paper, \acrshort{mlops} for sparse image
analysis will be explored. We propose a long-term vision holistic approach to enhance biomedical image analysis that includes: a fingerprinting process that enables selecting the best models, datasets, and model development strategy relative to the task at hand, therefore making use of available models and datasets; an automated model development stage; and a continuous deployment and monitoring process.

Our work is organized in the following manner. Section \ref{sec:related_work} explains the fundamental concepts related to our work and provides an overview of other related works, and Section \ref{sec:methodology} details the proposed approach. In Section \ref{sec:experiments}, the preliminary experiments carried out during the investigations are described. Their results are presented and discussed in Section \ref{sec:results}. Section \ref{sec:conclusion} summarizes our investigation and outlines future research directions.

\section{Background and Related Work}
\label{sec:related_work}
In this section, we provide fundamental notions around \acrshort{mlops} and other important fields related to our work. We also present state-of-the-art approaches related to ours in general and regarding the different building blocks in particular. 

\subsection{\acrshort{mlops}}
\label{subsec:mlops}
\begin{figure}[tb]
	\centering
	\includegraphics[width=0.75\textwidth]{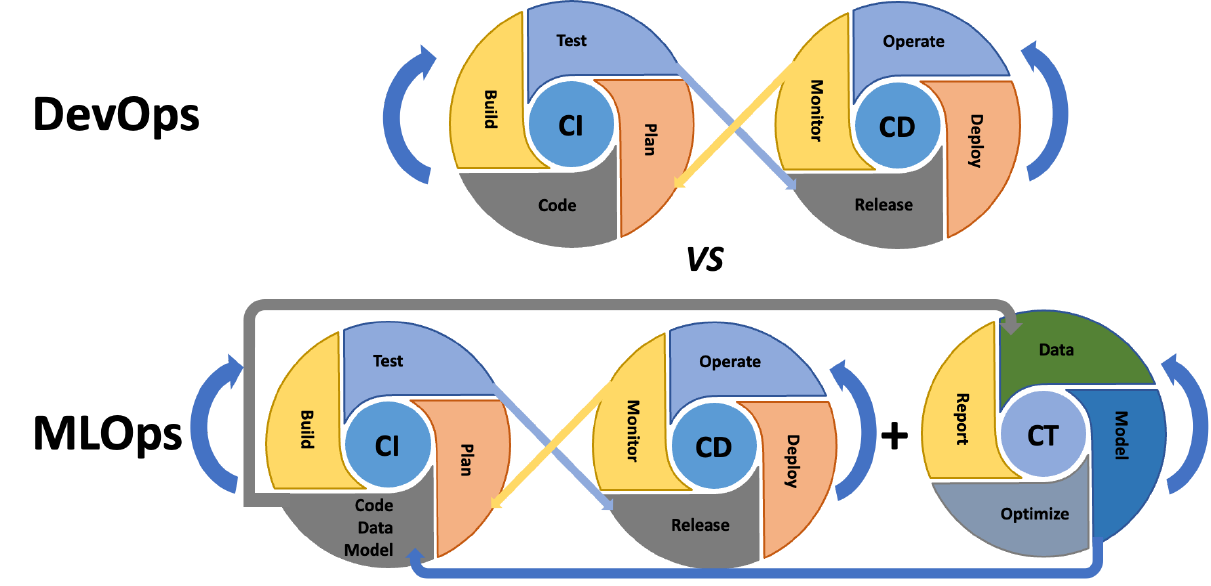}
	\caption{\acrshort{devops} vs. \acrshort{mlops} \cite{Islam_2023}. While \acrshort{devops} entails applying the main concepts of \acrfull{ci} and \acrfull{cd} on code only, \acrshort{mlops} has the two new components data and model, which are added to the code component. Additionally, \acrshort{mlops} has a new concept named \acrfull{ct}.}
	\label{fig:myfigure1}
\end{figure}

\acrshort{mlops} is a discipline that combines \acrshort{ml} with software engineering paradigms such as \acrshort{devops} and data engineering to enable efficient deployment and operationalization of \acrshort{ml} systems \cite{DBLP:journals/access/KreuzbergerKH23, DBLP:conf/ccwc/SymeonidisNKP22}. It can be seen to a certain extent as \acrshort{devops} for \acrshort{ml} systems. The key differences and similarities between \acrshort{devops} and \acrshort{mlops} can be seen in Figure \ref{fig:myfigure1}. On the one hand, both entail two main concepts \acrlong{ci} and \acrlong{cd}. 
\begin{itemize}
\item \acrfull{ci}: consists in automatically building, testing and validating source code.
\item \acrfull{cd}: enables frequent release cycles by automatically deploying the software in production. It distinguishes itself from continuous delivery by automatizing the deployment process.
\end{itemize}
On the other hand, in addition to normal source code, \acrshort{ml} systems bring two new components: data and model, which are code-independent and therefore maintained separately from the code. This leads to the creation of pipelines to perform the complete processing. As a result, a novel concept specific to \acrshort{mlops} named \acrfull{ct} emerges, which involves automatically retraining and serving the models. Furthermore, the \acrshort{ci} and \acrshort{cd} concepts in \acrshort{mlops} differ from those in \acrshort{devops} in that \acrshort{ci} does not only include integrating new code and components but also new data, models and pipelines, and \acrshort{cd} does not deploy a single software package, but a complete \acrshort{ml} training and/or serving pipeline. Additionally, continuous monitoring,  i.e., automatically monitoring the IT system to detect potential problems such as compliance issues and security threads and address them, is extended in \acrshort{mlops} by monitoring production data and model performance metrics. This enables real-time understanding of model performances \cite{DBLP:journals/access/TestiBFIMSV22}.

Although \acrshort{mlops} is a relatively new field, important work and progress have been made. While some researchers focus on properly defining \acrshort{mlops} and providing an overview of its concepts and best practices \cite{DBLP:journals/access/KreuzbergerKH23, DBLP:conf/ccwc/SymeonidisNKP22, DBLP:journals/access/TestiBFIMSV22}, others investigate the challenges faced in \acrshort{ml} systems operationalization. Tamburri highlights in \cite{DBLP:conf/synasc/Tamburri20} the limited attention given to \acrshort{mlops} within academia  and the lack of data engineering skills in academia as well as in industry. Renggli \etal \cite{DBLP:journals/debu/RenggliRGK0021} and Granlund \etal \cite{DBLP:conf/icse-wain/GranlundKSMM21} extend on this and show that one massive problem in \acrshort{mlops} is data management. This is mostly due to the strong dependency between model performance and data quality. The cloud architecture center of Google proposes in \cite{mlopsgoogle} how to automate \acrshort{ml} workflows. They classify \acrshort{mlops} into three different levels: \acrshort{mlops} Level 0 that refers to classical \acrshort{ml} pipelines with no \acrshort{ci}, no \acrshort{cd} and manually operated workflows totally disconnected from the \acrshort{ml} system, \acrshort{mlops} Level 1 in which data validation, model validation, continuous delivery and \acrshort{ct} are introduced, and \acrshort{mlops} Level 2 where \acrshort{ci}, \acrshort{ct} and continuous delivery are fully explored. They apply semi-automated deployment in a pre-production environment and manual deployment in the production environment.

Several \acrshort{mlops} tools have been developed. A non-exhaustive list of tools and their features can be found in \cite{DBLP:journals/access/KreuzbergerKH23, DBLP:conf/euromicro/RecupitoPCMNPT22, DBLP:conf/ccwc/SymeonidisNKP22, DBLP:journals/access/TestiBFIMSV22}. There is no ideal tool, as each tool covers different \acrshort{mlops} aspects. In practice, tools are often combined to achieve maximal efficiency. However, the majority of tools focus on model versioning and tracking and ignore dataset versioning. This impedes the ability to reproduce results and renders it reliant on the coding practices of skilled experts \cite{DBLP:conf/icsa/ZarateMDT22}. In industry particularly, due to IT-Software's large and complex nature, the \acrshort{mlops} tools are often diverse and must match a specific established strategy.

Despite advancements in the \acrshort{mlops} domain, there are very few published real-world use cases in which \acrshort{mlops} is clearly designed, explained and applied. One use case found is Oravizio \cite{Granlund_Stirbu_Mikkonen_2021}, a medical software for evaluating the risks associated with joint replacement surgeries. The researchers had four different risk models from which the best was selected and deployed. One issue they faced during development was related to data management, due to the multitude of data formats they had to process. A second use case is SemML \cite{DBLP:journals/ws/ZhouSGSCMWK21} in which the researchers propose a \acrshort{ml}-based system that leverages semantic technologies to enhance industrial condition monitoring for electric resistance welding. Their workflow enables them to reuse and enhance \acrshort{ml} pipelines over time based on new input data. A third example is microbeSEG \cite{10.1371/journal.pone.0277601}, a \acrshort{dl}-based tool for instance segmentation of objects. The researchers automate several data management tasks and model building processes. They, however, do not focus on monitoring and deployment. 

We found two ongoing works connected to ours. Firstly, Friederich \etal explore in \cite{nils} \acrfull{ai} to encode dynamic processes. They propose a \acrshort{mlops}-based pipeline, in which active learning will be used to predict and record potentially important events during experiments in light microscopy. Secondly, Z{\'{a}}rate \etal present  K2E \cite{DBLP:conf/icsa/ZarateMDT22}, a new approach to governing data and models. They investigate \acrshort{mlops} environments for creating, versioning and tracking datasets as well as models. They are still in the conceptualization step and plan to build their platform by following the \acrfull{iac} paradigm. To the best of our knowledge, there is no \acrshort{mlops} approach similar to ours, particularly in the field of biomedical image analysis.
\subsection{\acrshort{automl} and Meta-learning}
\label{subsec:automl}
\acrfull{automl} is the process of automating different stages of the \acrshort{ml} development cycle. These stages often include data preprocessing, feature engineering, model training and model evaluation. \acrshort{automl} addresses problems such as \acrfull{dac}, \acrfull{hpo}, \acrfull{cash} and \acrfull{nas} \cite{10.1145/3470918}. Several researchers work towards building fully automated systems for their applications. For example, Meisenbacher \etal explain in \cite{meisenbacher2021concepts} how to achieve \acrshort{automl} for forecasting applications. They define five levels of automation for designing and operating forecasting models. \\Numerous tools have been developed, each addressing specific \acrshort{automl} sub-problems. The developed tools often log metadata such as hyperparameters tried, pipelines configurations set, model evaluation results, learned weights and network architectures. Based on these experiences and other dataset metadata, a model is trained with the aim to adapt faster to new tasks \cite{DBLP:journals/pami/HospedalesAMS22}. This is meta-learning, also known as learning to learn.

While it is clear that \acrshort{automl} can be combined with \acrshort{mlops} to enable high level automation in the \acrshort{ml} development lifecycle \cite{mlopsgoogle, DBLP:conf/ccwc/SymeonidisNKP22}, exploring the output of this combination for meta-learning seems not to be investigated. The information acquired during the continuous monitoring stages appears not to be widely researched in combination with \acrshort{automl} or meta-learning.
\subsection{Scarce Image Data}
\label{subsec:scarce_image_data}
Scarce image data is a massive problem in \acrshort{ml}. In the field of image analysis, particularly in biomedicine, the acquisition and annotation of images must be done by highly skilled experts, require a considerable amount of time and resources, and are often error-prone \cite{DBLP:journals/jib/SchillingSKR22}. As a result of these constraints, the amount of image data present is either small in quantity, dominated by noise, or small in quality, very weakly or not labeled. A lot of approaches have been developed to address data scarcity. On the one hand, there are image processing techniques to augment image data such as scaling, rotation and cropping. On the other hand, \acrshort{dl} methods such as \acrfull{gan} \cite{goodfellow2020generative} and \acrfull{vae} \cite{DBLP:journals/corr/KingmaW13} are used to generate synthetic images. These \acrshort{dl} approaches often use \acrfull{ssl} to understand better how data points are sampled \cite{marcel}.  We notice however that these solutions often focus more on image classification tasks.
\subsection{Image Fingerprinting}
\label{subsec:image_fingerprint}
Fingerprinting is used in image processing to generate concise and distinct representations of images. It is useful for diverse objectives such as image retrieval, copyright protection, or image similarity analysis. We focus on image fingerprinting to measure the similarity between images and/or datasets.

Ranging from simple pixel distribution methods \cite{Mitchell2010} to \acrshort{dl} approaches \cite{https://doi.org/10.1002/aic.690370209}, there are numerous methods to compute similarity between images. Godau and Maier-Hein present in \cite{10.1007/978-3-030-87202-1_42} an image fingerprinting approach that consists of embedding images along with their labels in a fixed-length vector in order to capture semantic similarities in biomedical image datasets. Molina-Moreno \etal \cite{marcel} build an \acrfull{ae}, train this using \acrshort{ssl} and obtain a two-dimensional latent space in which the disposal of image datasets displays their similarity. Such similarity measures enable researchers to apply transfer learning for their respective tasks. This is often achieved by selecting suitable pre-trained models and/or datasets for a new task based on the similarity measure obtained.  Nevertheless, most state-of-the-art methods compute this similarity on a dataset level or on an image level and do not investigate the computation on an image patch level.

To fill the observed gaps in the context of image analysis, we propose a new approach, which is fully described in the following section. 

\section{Methodology}
\label{sec:methodology}
This section describes the proposed methodology to address the different problems identified and mentioned in Section \ref{sec:introduction} and Section \ref{sec:related_work}. We first provide a global description of the system and subsequently delve into its individual building blocks.
\subsection{Overview of the proposed approach}
\label{sec:methodology_1}
\begin{figure}[tb]
	\centering
	\includegraphics[width=1\textwidth]{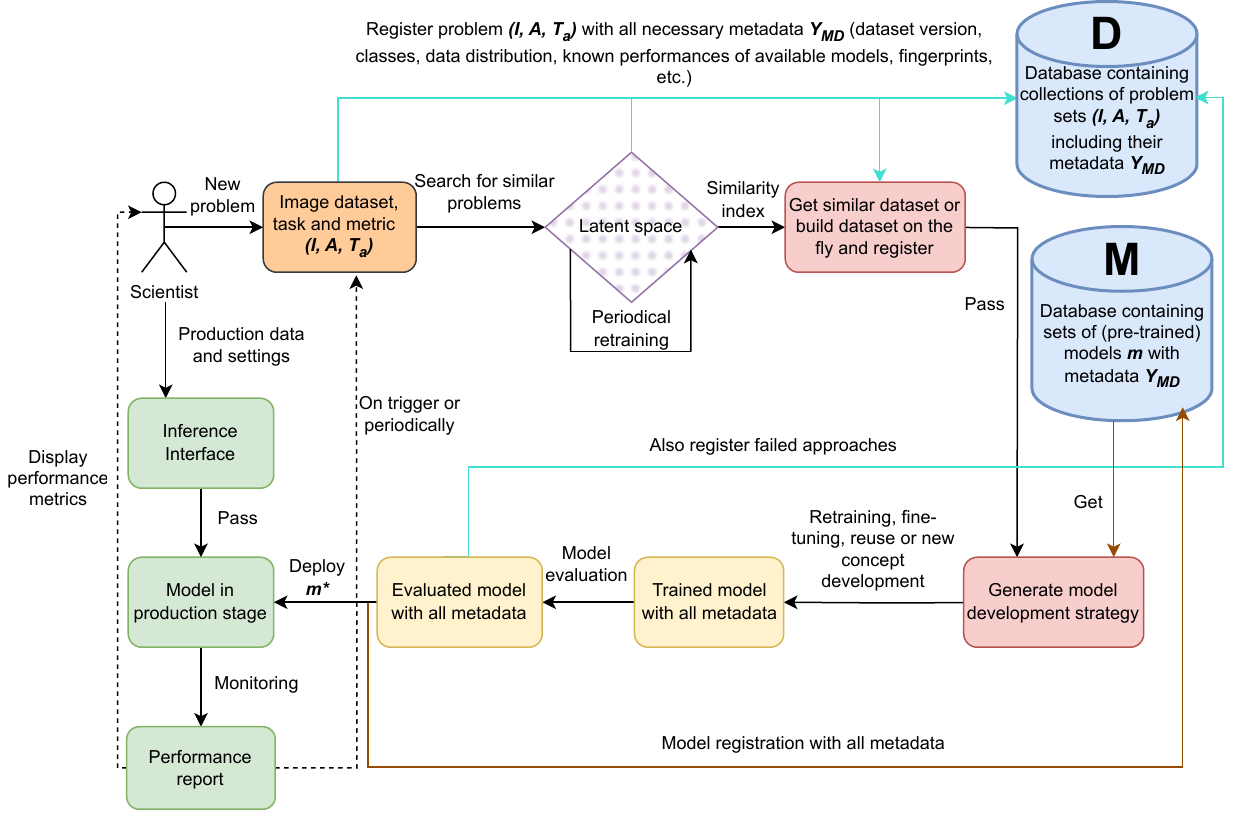}
	\caption{Abstract architecture of the proposed \acrshort{mlops}-based image analysis approach. The turquoise arrows indicate exchanges with the data database $\mathbb{D}$ and the brown arrows exchanges with the model database $\mathbb{M}$. The black arrows model the information flow between different building boxes of the system and the dashed lines feedback from production to development and to the scientist. The orange box represents the input provided by the scientist. The red boxes are the meta-learning system that handles dataset and algorithm selection. The yellow boxes display the \acrshort{automl} pipeline. The green boxes represent the continuous deployment and monitoring stage for production.}
	\label{fig:myfigure2}
\end{figure}

Figure \ref{fig:myfigure2} shows a conceptual architecture of our method. 
The main entering point is a scientist bringing a new image analysis task modeled as the triple$(I, A, T_a)$, where $I$ represents the image dataset, $A$ the performance analysis metric, e.g., F1-Score, and $T_a$ the task, e.g., image classification. At a time $t$, the performance metric $A_t$ can be computed as defined in Equation \eqref{eq:one}, in which $m_t$ represents the current model, $I_t$ the current image dataset and $Y_{MD,t}$ the metadata relative to $I_t$. The following stages of our approach attempt to find the best model $m^*$ defined through Equation \eqref{eq:two}.
\begin{equation}
	A_t = m_t(T_a, I_t, Y_{MD,t})_{m \in \mathbb{M}}
	\label{eq:one}
\end{equation}
\begin{equation}
	m^*(t) = \underset{m \in \mathbb{M}}{\arg\max} \ (A_t)
	\label{eq:two}
\end{equation}

\subsection{Image Similarity}
At $t=0$, the newly provided problem is initialized and registered in the image database $\mathbb{D}$. The initial image dataset $I_0$ along with the task $T_a$, the performance analysis metric $A_0$ set to $-\infty$ and other metadata $Y_{MD,0}$ such as dataset version, classes, data distribution, fingerprints, known performances of available models for the same task $T_a$, etc.\ build an entry in $\mathbb{D}$.  In order to discover images similar to $I_0$, its fingerprints $f_i(I_0)$ are subsequently computed, saved and compared to those already present in $\mathbb{D}$ through a latent space embedding built in advance. This embedding will be periodically retrained to verify that the performance meets expectations continuously.

As emphasized in \cite{10.1007/978-3-030-87202-1_42, marcel}, with such fingerprints, deploying new \acrshort{ml} models for biomedical applications can be seed up by selecting appropriate pre-training models. Furthermore, this could solve scarcity issues by finding appropriate images for image augmentation on the one hand and for pre-training on the other hand.

\subsection{Model Development Strategy}
Inspired by early attempts in \cite{DBLP:conf/miccai/YuanMJLCRPDN20} and \cite{DBLP:journals/tmi/ZhangWYSHTWRMXX20}, the model development strategy aims to leverage the concept of meta-learning for scarce data. Given the tuple $(I, A, T_a)$, the metadata $Y_{MD}$ acquired during the registration phase and the computed fingerprints $f_i(I)$, the goal is to find the top $k(k \geq 1)$ cheapest and efficient model developing approaches. These approaches include both model and dataset and could range from model selection together with potential weights and development strategy, i.e., fine-tuning, retraining, to dataset selection or conception on the fly. Dataset conception in this context involves selecting the closest images to $I$ in $\mathbb{D}$ and using these as training data.

The reason we envision using not only $f_i(I)$ is to minimize error propagation when fingerprinting is inaccurate. In this case, the meta-learner will solely rely on the metadata $Y_{MD}$.
\subsection{Automatic Model Development}
This stage is a direct application of the strategy established previously. It performs \acrshort{automl} according to the strategy defined. Although \acrshort{automl} is more computationally expensive than normal \acrshort{ml} techniques, it is more efficient and faster in producing the best-trained model \cite{mlopsgoogle, Microsoft}. It can be combined to \acrshort{mlops} to improve a project's automation level, therefore reducing the pipeline configurations overhead. For example, in a retraining process or new training, it could help solve the \acrshort{hpo} problem faster at time $t$.

All development runs, including failing approaches, will be tracked and recorded in the model database  $\mathbb{M}$. They would serve as input data for the meta-learner in the previous stage and may help identify flaws in the data or in the whole system. The best model $m^*$ is sent in the last stage.

\subsection{Continuous Deployment and Monitoring}

The best model developed $m^*$ is continuously deployed as a service and monitored during production. On the one side, we envision a deployment framework fulfilling fundamental requirements such as independency towards the \acrshort{ml} Framework, rapid maintenance, accessibility, and parallel computing. Despite the existence of numerous deployment frameworks, we will focus on \acrfull{rest} frameworks such as DEEPaaS \cite{García2019} and EasyMLServe \cite{neumann2022easymlserve}. On the other side, the performance metric $A_t$ as well as defined metrics by the scientist will be continuously monitored over time and reported. This monitoring will be particularly investigated, as it could help identify potential concept drift and decrease in performances, and act accordingly by triggering the complete framework from the start.

The high information reuse, which will be investigated in this work, will serve the purpose of exploiting available feedback and enabling transfer learning. Our approach will mostly be done using the Python programming language, due to its rich ecosystem of data science libraries and its expansive and highly engaged user community. We also plan to apply containerization with Docker \cite{10.5555/2600239.2600241} and its microservices, as it guarantees platform independency and enforces reproducibility.

\section{Preliminary Experiments}
\label{sec:experiments}
This section describes the current state of the investigation. The experiments performed mainly focus on the first stage of our approach described in Section \ref{sec:methodology} which consists in building a latent space embedding in which image data can be represented along with their similarities. To this end, we build an autoencoder whose architecture will be depicted in Section \ref{autoencoder}, and evaluate it on biomedical image datasets presented in Section \ref{dataset}. Our implementation can be found in \cite{pomlia}.
\subsection{Autoencoder}
\label{autoencoder}
An autoencoder is a special type of neural network that takes an input $x$, transforms it in a compressed and informative representation $x_{ae}$ and reconstructs the initial input based on $x_{ae}$. The goal is to find an encoding function $f: x \mapsto x_{ae}$, and a decoding function, $g: x_{ae} \mapsto \hat{x}$ such that the difference between the reconstructed input $\hat{x}$ and the initial input $x$ is minimal. This difference is measured by a loss $L$ : $\min_{f, g} L(x, g(f(x)))$. The goal is to obtain a powerful representation $x_{ae}$ that can be used for various tasks.

\begin{figure}[tp]
	\centering
	\includegraphics[width=0.9\textwidth]{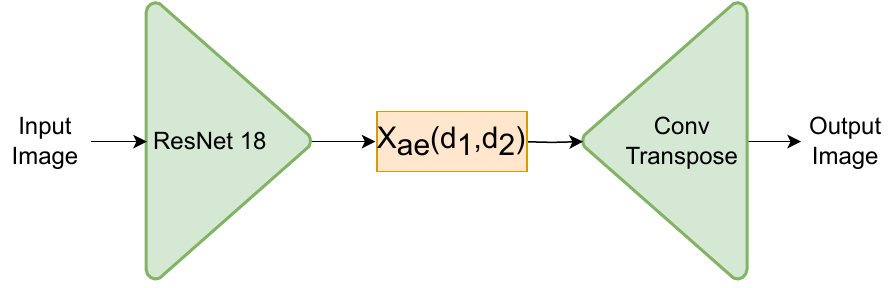}
	\caption{Autoencoder. The encoder is based on the ResNet18 architecture, the latent space representation is a two-dimensional space, and the decoder entails stack transpose convolutional layers.}
	\label{fig:figure3}
 \end{figure}

The architecture of our autoencoder can be seen in Figure \ref{fig:figure3}. Our encoder is a vanilla (standard) ResNet18 \cite{DBLP:conf/cvpr/HeZRS16} based neural network. The final fully connected layer is replaced by a linear layer mapping the 512 feature channels vector to a two-dimensional vector. Because we are more interested in $x_{ae}$, we build a simple decoder that entails five stacked convolutional transpose layers. We choose the \acrfull{mse} as loss function that computes the difference between the original input image and the reconstructed output image, and Adam as optimizer.
\subsection{Datasets}
\label{dataset}
\begin{figure}[tb]
	\centering
	\includegraphics[width=1\textwidth]{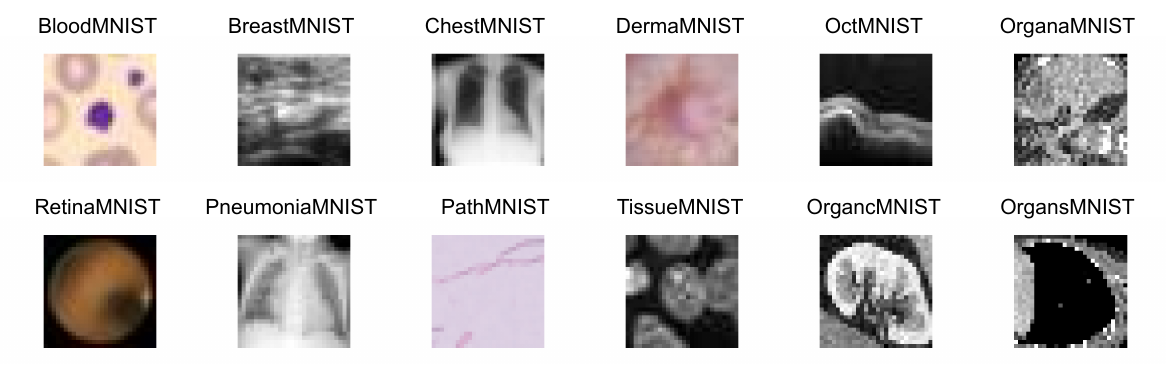}
	\caption{MedMNIST v2 2D datasets}
	\label{fig:myfigure4}
\end{figure}

To evaluate our autoencoder, we select all the 2D image datasets of the MedMNIST v2 \cite{DBLP:journals/corr/abs-2110-14795} dataset, a benchmark dataset for 2D and
3D biomedical image classification. The 12 selected datasets can be found in Figure \ref{fig:myfigure4}. They consist of eight gray-scale image datasets (BreastMNIST, ChestMNIST, OctMNIST, OrganaMNIST, OrgancMNIST, OrgansMNIST, PneumoniaMNIST and TissueMNIST) and four color image datasets (BloodMNIST, DermaMNIST, RetinaMNIST and PathMNIST).
The autoencoder is trained, validated and tested on the respective datasets splits. All the images are processed as $3 \times 32 \times 32$ images. For our architecture to be usable on all these datasets, the gray-scale images were loaded and processed as three-channel images.

\section{Results and Discussions}
\label{sec:results}
We present in this section the results of the experiments performed in the previous section and discuss these.

\begin{figure}[tb]
	\centering
	\includegraphics[width=0.9\textwidth]{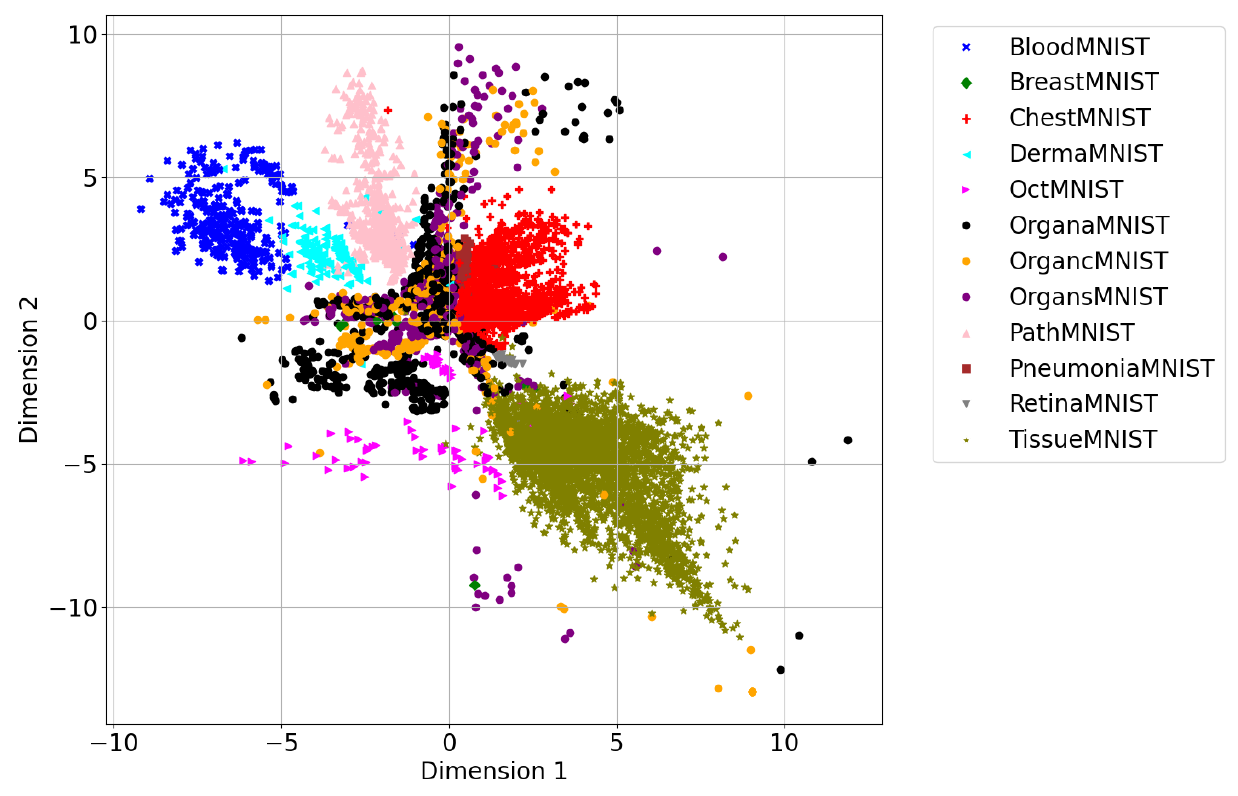}
	\caption{Latent space representation of the MedMNIST v2 2D datasets (test sets)}
	\label{fig:myfigure5}
\end{figure}

Figure \ref{fig:myfigure5} shows the latent space representation of $N=10000$ test samples collected from the 12 2D image datasets presented in the previous section. We notice that the color image datasets BloodMNIST, DermaMNIST and PathMNIST build a cluster at the top left. This is most likely due to the close distribution of the pixel values of the respective images. These color image datasets are, however dissimilar to the color image dataset RetinaMNIST, in which the images of the retina all have a black contour.
 The OrganaMNIST, OrgancMNIST and OrgansMNIST datasets are mixed and almost not differentiable, leading to a non-identification of particular structures in the latent space. This is because all three datasets have images of the same organ taken on different views. In OrganaMNIST, the images are acquired on an axial view, in OrgancMNIST in a coronal, and in OrgansMNIST in a sagittal view. Nevertheless, all three datasets have multiple outliers that may hinder the model in positioning new incoming images.

\begin{figure}[tb]
	\centering
	\includegraphics[width=0.9\textwidth]{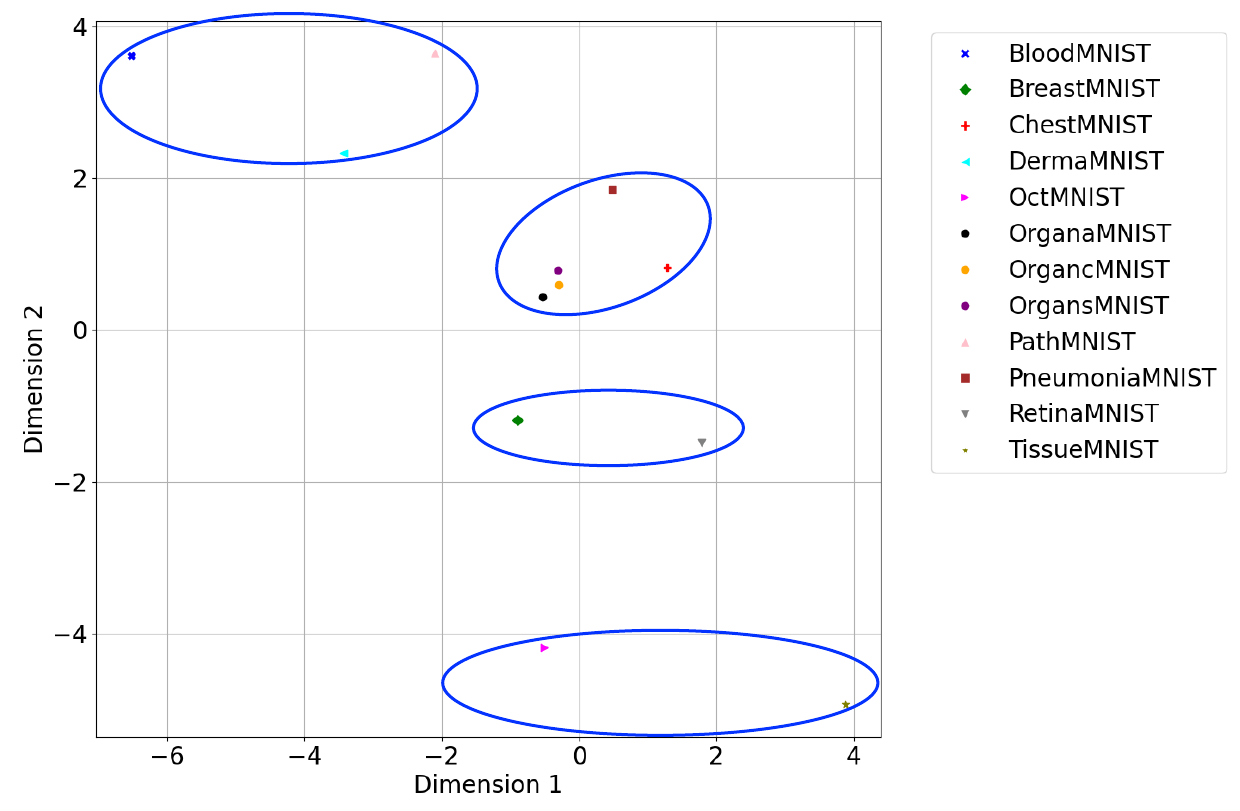}
	\caption{Mean value latent space representation of the MedMNIST v2 2D datasets (test sets)}
	\label{fig:myfigure6}
\end{figure}

A better view of the latent space is provided in Figure \ref{fig:myfigure6}, in which the mean of all embedding vectors of the test images are computed for each dataset. We notice four main clusters. The first cluster entails the three-channel image datasets BloodMNIST, DermaMNIST and PathMNIST, the second cluster in which OrganaMNIST, OrgancMNIST and OrgansMNIST are very close, as well as PneumoniaMNIST and ChestMNIST. The third cluster consists of BreastMNIST and RetinaMNIST, and the final cluster of TissueMNIST and OctMNIST.

This autoencoder shows a promising ability to capture similarities between images. Despite being in the early stages of our investigation, we strongly believe that this approach to solving the image similarity question is encouraging and could be fine-tuned to identify specific structures in image crops.

\section{Conclusion and Future Work}
\label{sec:conclusion}
In this paper, we presented a new approach to improve biomedical image analysis. Our approach aimed to apply \acrshort{mlops} to image analysis tasks, particularly when the image dataset is scarce. To achieve this goal, we presented a multi-stage framework that leverages the existence of models and benchmark datasets to solve a given task. The first stage enables us to find image datasets similar to the images of the given task. The second stage applies meta-learning to select the best model development strategy for the given task. This strategy is executed via the third stage of \acrshort{automl}. The final stage deploys the best-trained model and monitors the model's performance continuously to achieve optimal performance.

The preliminary experiments carried out in this paper mostly focused on the first stage, which consists in computing image fingerprints to identify similar datasets. We built a ResNet18-based autoencoder and showed that the resulting 2D latent space representation is interpretable enough to find similarities between images. We, however, faced some challenges when the images are all from the same object but taken from different angles and focused only on 2D image datasets.

Our future research will, therefore, focus on extending the image fingerprinting process to 3D image datasets and even to the level of image patches and improving the representation of different artifacts. Understanding the image at this granularity level would enable us to generate data or labels to solve the data scarcity problem. We also plan to investigate the effects of outliers, as they may highly impact the similarity measurements. Finally, we will continue our research on the other stages of the proposed approach, including meta-learning for biomedical image analysis, \acrshort{automl} and efficient model deployment and monitoring.

\section*{Acknowledgements}
\label{sec:acknowledgements}
This project is funded by the Helmholtz Association under the program "\acrfull{nacip}". We would like to thank the \acrfull{hai} for providing the computing infrastructure via the \acrfull{haicore}. We also extend our gratitude to all those who provided assistance, even if not explicitly mentioned here.

The authors have accepted responsibility for the entire content of this manuscript and approved its submission. We describe the individual contributions of A. Yamachui Sitcheu (AYS), N. Friederich (NF), S. Baeuerle (SB), O. Neumann (ON), M. Reischl (MR), R. Mikut (RM): Conceptualization: AYS, NF, RM; Methodology: AYS, NF; Software: AYS, SB; Investigation: AYS, NF, SB; Writing – Original Draft: AYS; Writing – Review \& Editing: AYS, NF, SB, ON, MR, RM; Supervision: RM; Project administration: RM; Funding Acquisition: MR, RM.

\addtocontents{toc}{\protect\newpage}



\end{document}